\documentclass[conference]{IEEEtran}

\ifCLASSOPTIONcompsoc
  \usepackage[nocompress]{cite}
\else
  \usepackage{cite}
\fi
\ifCLASSINFOpdf
   \usepackage[pdftex]{graphicx}
   \graphicspath{{./pdf/}{./image/}}
   \DeclareGraphicsExtensions{.pdf,.jpeg,.png}
   \usepackage{lipsum}
\else
\fi
\begin{document}
\title{A Multi-Scale Cascade Fully Convolutional Network Face Detector}

\author{\IEEEauthorblockN{Zhenheng Yang, Ramakant Nevatia}
\IEEEauthorblockA{Institute for Robotics and Intelligent Systems\\University of Southern California\\
Los Angeles, California 90089\\
Email: \big\{zhenheny,nevatia\big\}@usc.edu}}
 
% make the title area
\maketitle

% As a general rule, do not put math, special symbols or citations
% in the abstract
\begin{abstract}
Face detection is challenging as faces in images could be present at arbitrary locations and in different scales. We propose a three-stage cascade structure based on fully convolutional neural networks (FCNs). It first proposes the approximate locations where the faces may be, then aims to find the accurate location by zooming on to the faces. Each level of the FCN cascade is a multi-scale fully-convolutional network, which generates scores at different locations and in different scales. A score map is generated after each FCN stage. Probable regions of face are selected and fed to the next stage. The number of proposals is decreased after each level, and the areas of regions are decreased to more precisely fit the face. Compared to passing proposals directly between stages, passing probable regions can decrease the number of proposals and reduce the cases where first stage doesn't propose good bounding boxes. We show that by using FCN and score map, the FCN cascade face detector can achieve strong performance on public datasets.
\end{abstract}

% no keywords

% For peer review papers, you can put extra information on the cover
% page as needed:
% \ifCLASSOPTIONpeerreview
% \begin{center} \bfseries EDICS Category: 3-BBND \end{center}
% \fi
%
% For peerreview papers, this IEEEtran command inserts a page break and
% creates the second title. It will be ignored for other modes.
\IEEEpeerreviewmaketitle

\section{Introduction}
Face detection is a widely-studied but still challenging problem in computer vision. Modern face detectors can reliably detect near-frontal faces, but challenges still exist when the images are taken "in the wild". We focus on the problem of face detection in a single image or a single frame in a video.

The difficulties of face detection come largely from two causes: 1) large variations due to illumination, expression, occlusion and poses; 2) large search space as faces can be at arbitrary location and of any size. The first difficulty requires the face detector to be robust against the variations, while the second requires fast binary classification.  We propose a fully convolutional cascade network to meet these challenges. 

Modern face detector approaches can be categorized along three different axes. One is the types of features used which have ranged from simple Haar-like features in the seminal work of Viola and Jones \cite{viola2005detecting} to SURF features in \cite{li2013learning} and convolutional nearual network (CNN) features in detectors such as in \cite{vaillant1994original}. Another important dimension is the type of classifier used: while various classifiers such as SVMs have been used, cascade classifiers have been popular due to their efficiency \cite{viola2005detecting, li2013learning, chen2014joint}, \cite{li2015convolutional}, \cite{mathias2014face}. Finally, methods vary based on whether the computed features are aggregated over an entire region or a part-based analysis is performed; this set includes Deformable Part Model \cite{mathias2014face}, Tree Parts Model (TSM) \cite{zhu2012face}, structure model \cite{yan2014face}, Deep Pyramid Deformable Part Model \cite{ranjan2015deep}, Faceness \cite{yang2015facial}.

In our approach we adopt modified CNN features and cascade classifier; we do not use a part-based analysis but such reasoning can be added to our framework. Our work is inspired by recent work of Li et al \cite{li2015convolutional} which proposed a CNN cascade framework consisting of three levels. The first level of the CNN cascade generates fixed size proposals which are then resized and passed on to the next level. The scanning stride of proposal windows is determined by the network architecture. The parameters should be optimized carefully for accurate proposal generation. Reference \cite{li2015convolutional} showed good performance on a number of datasets, exceeded only by part-based methods such as Faceness, which are computationally much more expensive.

We make a critical change from approach of \cite{li2015convolutional} by introducing a multi-scale fully convolutional network (FCN) \cite{jonathanlongfcn} and a score map. FCN replaces the fully connected layers in a CNN architecture with a fully convolution layer and a score map is used for face localization. Instead of a label or a feature vector, the output of an FCN is termed a \textit{heatmap}. Each point in the heatmap corresponds to a region in the original image; the corresponding activation of the region is used as the proposal probability. Score map is generated by adding up the heatmaps under different scales. Thus every point in the original image has a probability of being a face. This probability is then used to localize faces. Fig. 1 is the flowchart showing how the multi-scale FCNs generate a score map and its use in generating face proposals which are verified and located accurately. 

Use of FCNs allows us to create proposals of different scales and removes the need for resizing the proposals for the following stages. Use of score map also makes the detector robust to noise in training samples. In most face datasets, rectangle annotations around the faces inevitably contain part of the background which adds noise during training.

Our method shows state-of-art performance on some public datasets and overall performs comparably to leading face detectors, such as DPM and Faceness but at a significantly lower computational cost.

We make the following contributions in the paper:
\begin{itemize}
\item We developed an FCN cascade that gradually zooms in to the faces and introduce the training for the network.
\item We demonstrated a multi-stream strucuture to deal with multi-scale face detection.
\begin{figure*}
\label{fig:1}
\includegraphics[width=\textwidth, height = 68mm]{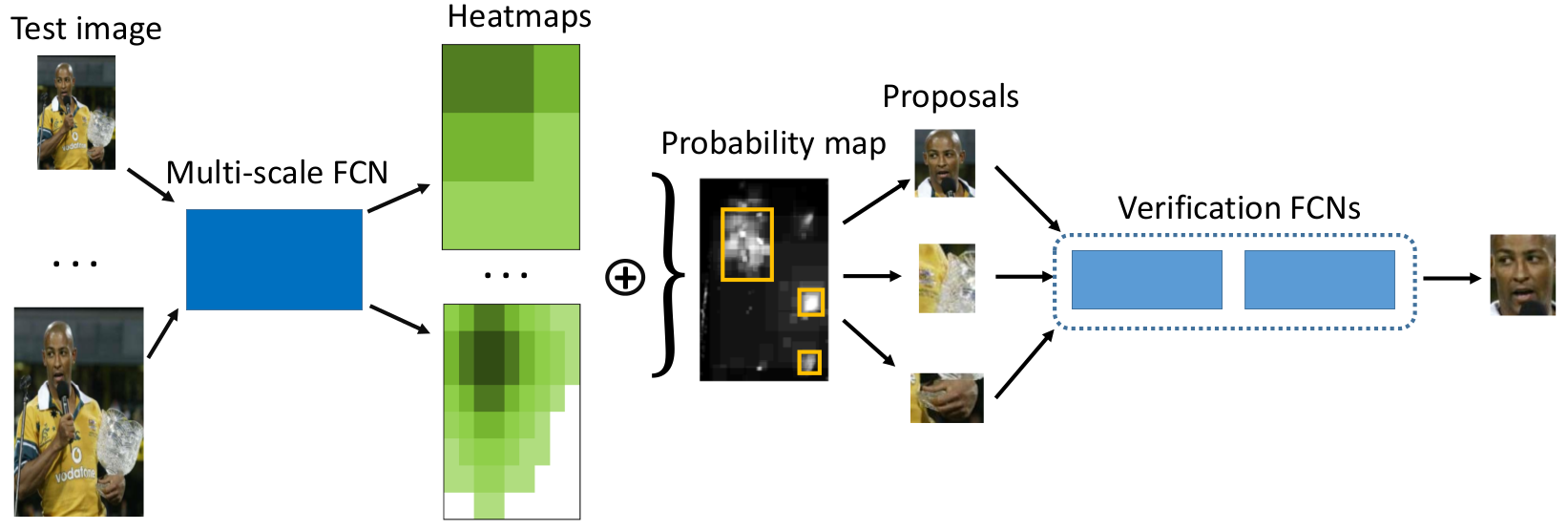}
\centering
\caption{Illustration of the FCN cascade framework. Given a test image, it's scaled and fed into multi-scale FCN. Score map is generated from FCN and proposals are picked from the score map. The proposals are then fed into the verification FCNs, which are trained on hard  examples.}
\end{figure*}

\item We show that by using FCN and score map, our method achieves state-of-art performance under some cases.
\end{itemize}

\section{FCN Cascade Framework}

The FCN cascade framework is composed of three stages of an FCN. For an test image, the first stage of the FCN generates a list of potential boxes from a score map. The later two stages gradually zoom in on the boxes to verify and more accurately localize faces by generating score map on these boxes from the previous stage. Fig. 1 illustrates the overall FCN cascade framework.

\subsection{FCN Overview}
CNN has achieved success on image classification \cite{krizhevsky2012imagenet}, face detection \cite{li2015convolutional}, attribute leaning \cite{gan2016learning} and video classification \cite{gan2015devnet}. The idea of using an FCN is to extend a CNN to arbitrary-sized inputs. Each layer of data in this network is a three dimensional array. In the first layer, the size of the array corresponds to the size of  the input image and channel number. Each point in higher level of data corresponds to some region in input image. This region is called its \textit{receptive field}.

A standard CNN (e.g. AlexNet \cite{krizhevsky2012imagenet}) has a few fully connected layers (\textit{fc} layers) as the last layers. Neurons in these layers have full connections with neurons in the previous layer. For detection task, the \textit{fc} layers serve as binary classifier which can only take fixed size inputs. For an input image of size \begin{math} W\times H \end{math} with three color channels, convolutional layers generate a feature map of size  \begin{math}W'\times H' \times D.\end{math} \begin{math}D\end{math} is the output feature dimension. \begin{math}W'\end{math} and \begin{math}H'\end{math} is determined by the input image size, convolutional kernel size, convolutional stride, pooling size,  pooling stride and padding size. When the network architecture is fixed, the input image size has to be fixed to fit the \textit{fc} layers.

FCN replaces the \textit{fc} layer with a fully convolutional layer.
Writing $x_{ij}$ for the data vector in some layer, $y_{ij}$ for the data after a fully convolutional layer. The fully convolutional layer computes output $y_{ij}$ by

\begin{equation} \label{eq:1}
y_{ij} = f_{ks}(\{x_{si+s\delta}\}_{, 0<\delta<s})
\end{equation} 
where $s$ is the sampling stride of convolution, $k$ is the kernel size. $f_{ks}$ is determined by the type of layer. It is a matrix multiplication for convolutional layer or average pooling, a spatial max for max pooling, or an elementwise nonlinearity for ReLU layer.  The sampling stride of the whole network is determined by the network architecture.
\begin{equation} \label{eq:2}
S = \prod_i s_i \cdot \prod_j k_j
\end{equation} 
in which, $s_i$ is the stride in convolutional layer $i$, $k_j$ is the kernel size in pooling layer. The output of the network is termed a \textit{heatmap}. Each element in the heatmap corresponds to its receptive field. Compared with region sampling with sliding windows or bounding box proposals, the structure of FCN makes it possible for end-to-end training. It also allows feature sharing among boxes with overlapping regions, which significantly speeds up computation.

\subsection{Proposal Generation}
The first stage in our framework aims to generate face bounding box proposals with the score map of an FCN. For an input image $\Gamma$, we want to learn a score function:
\begin{equation} \label{eq:3}
P(R,\Gamma) \in [0,1]
\end{equation}
where $R = \{x,y,w,h\}$ is a rectangle box denoted by its up left corner coordinates, width and height. The value of $P$ denotes the probability that a face is contained in the box. 

\textbf{Multi-scale FCNs.} One issue with using an FCN in detection is that the receptive field size and stride are determined by the network architecture, while the size of the faces can vary significantly, even in the same image. Our solution is to use a multi-stream FCN. Assume that the FCN has been trained with inputs of the same scale. We extend this FCN into $n$ streams. For each stream, the FCN shares the same parameters as in the trained FCN. For testing an image, we resize the image into n different scales ($\Gamma_1$, $\Gamma_1$, … , $\Gamma_n$), and feed these different scaled images into the corresponding FCNs. The output heatmaps of each stream correspond to different sizes of receptive fields. 

\textbf{Score maps} $n$ streams of FCN generate $n$ heatmaps with each element corresponding to different scales in original image. We project the heatmaps back to the size of the original image and then add up the projected score maps to form a score map. Each heatmap can be projected back to the size of original image with pixel value in receptive field equal to corresponding score in heatmap.  Thus each element in the score map is a score for each pixel belonging to a face region. With this score map, we propose a list of boxes that potentially contain faces.

\begin{figure}[h]
\label{fig:2}
\includegraphics[width=0.48\textwidth, height = 63mm]{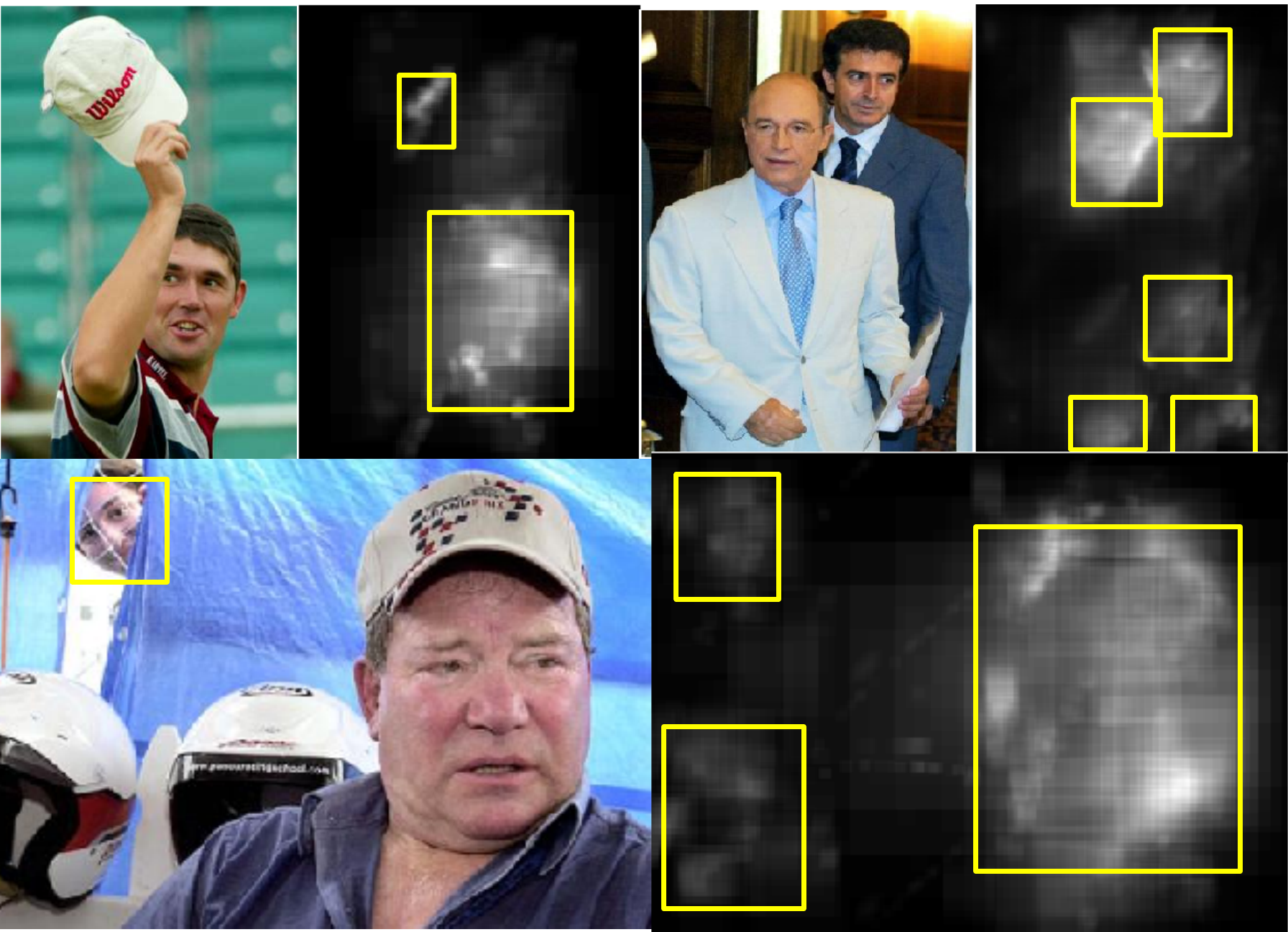}
\centering
\caption{Score maps are generated after each FCN. This figure shows the score maps generated from first stage of FCN.}
\end{figure}
 
	The proposals are generated by calculating a probability score $\Omega$ for all boxes and the boxes with higher than threshold $\Omega$ values are picked. The threshold $\Omega$ value is chosen empirically by experiments and kept fixed for all our tests.
\begin{equation} \label{eq:4}
\Omega = \sum(p_{ij})\times\Big(\frac{\sum(p_{ij})}{h\times w}\Big)_{i<h, j<w}
\end{equation}
in which $p_{ij}$ is the probability score in the box, $h$ and $w$ are the height and width of the box respectively. The score considers both the overall probability that the box contains a face and also that the box is not too large compared to the face region. The score can be efficiently calculated with an integral image. 

$\Omega$: bounding box with height of $h$ and width of $w$

$p_{ij}$: score for pixel $(i,j)$ within the bounding box.

The boxes picked from the score map will be larger than the boxes picked from each stream's heatmap. This strategy helps maintain a high recall rate in the first stage. More importantly, since the positive training samples inevitably contain some background, the trained FCN from these samples is not expected to be very accurate in localizing the face. Under some cases, the response can also be high on common background in the face region (\textit{e.g.} collar or hair) and thus the bounding boxes directly from the score map can be biased to these background.

\textbf{Training strategy.} The aim of the first stage of FCN cascade is to generate proposal boxes that contain a face. Thus the score function (as in (\ref{eq:1})) should be discriminative between the face and  background. We crop faces from the VGG Face Dataset \cite{parkhi2015deep} as positive samples and randomly crop patches from MIT Places Database for Scene Recognition \cite{zhou2014learning} from negative samples. Considering that for a real image, most of the boxes in the image don't contain a face, we set the negative/positive sample ratio to be high. So the trained network can be more efficient in rejecting non-face proposals in the first level. In our actual implementation, 6000 positive samples and 80,000 negative samples are used.

\textbf{Implementation.} Our n-stream FCNs are implemented in MatConvNet \cite{vedaldi2015matconvnet}. The first level of FCN has two convolutional layers and one fully convolutional layer.  After each convolutional layer, max pooling and ReLU are used as in AlexNet. Since this network is shallow, we can train it from scratch. The first convolutional layer has 3 input planes, 16 output planes and kernel size of 3. The second convolutional layer has 5 input planes and 16 output planes with kernel size of 5.  The sampling stride for such architecture is 2 and the window size is 30.
	We set the number of streams to be 6. Every test image is resized to be 600, 400, 260, 170, 100, 60 pixels. As the aspect ratio may change, the longer edge is resized to the length above.  Under this scheme, boxes of size from 30 pixel to 300 pixel in an image of longer edge equal to 600 pixel can be proposed to the next stage.
	
\subsection{Proposal Verification}
By setting a threshold, a small subset of boxes can be selected and passed to the later stages. The first stage serves to maintain a high recall rate while filtering out some easy negative boxes.

\textbf{Architecture of verification stages.} The verification stage in our FCN cascade is a series of two FCNs. In this stage, accuracy is more important than time efficiency, hence the two FCNs are designed to have a  more complex architecture than the first stage. We implement an FCN of 4 convolutional layers followed by an AlexNet with last layer replaced with a fully convolutional layer. The first FCN is trained from scratch with boosted training samples. The second FCN uses parameters from a pretrained model. We fine-tune the network by freezing earlier layers and just fine-tuning the last fully convolutional layers. 

\textbf{Training strategy.}  Ideally, we want the verification stage to handle hard examples. We mine hard negative and hard positive samples. When a proposed region from an image not containing a face has a high score, we take it to be a hard negative. Similarly, a proposed region containing faces but having a low score is treated as a hard positive. 

Intersection over union (IoU) value is used to determine whether a face is contained or not contained in a box. 
\begin{equation} \label{eq:5}
IoU = \frac{DT\cap GT}{DT\cup GT}
\end{equation}
$DT$ is the detection result rectangle, and $GT$ is the ground truth rectangle. Besides the hard negative and hard positive samples, we also add boxes that have a IoU values from 0.1 to 0.3 to the hard negative set. These help the verification stage learn to zoom in from large boxes around the faces to more accurate locations and scales of the boxes.

\textbf{Discussion.} Decomposing the process of detection to two stages of proposal generating and verification makes the network achieve high accuracy while maintaining high time efficiency as shown in the results below. The framework is also robust to noise in annotations.

\section{Experiments}
\textbf{Experimental setup.} We used four datasets to evaluate the performance of our framework. These are: Annotated Faces in the Wild (AFW) \cite{zhu2012face}, Face Detection Data Set and Benchmark (FDDB) \cite{jain2010fddb} PASCAL Face datasets \cite{yan2014face} and IARPA Janus Benchmark A (IJB-A) dataset \cite{klare2015pushing}. We compare our results with those of other leading methods; the methods compared with vary a bit across the datasets based on the availability of those results.

\textbf{AFW dataset results.} This dataset contains 205 images with 473 labeled faces from Flickr. We evaluate our detector on this dataset and compare with well known research and commercial face detectors. Research detectors include: \cite{zhu2012face}, \cite{yan2014face}, \cite{mathias2014face}, \cite{yang2015facial}, \cite{shen2013detecting}. Commercial detectors include \textit{Face.com}, \textit{Face++} and the best performing commercial face detector \textit{Google Picasa}. As can be observed from Fig. 3, our method is tied with DPM and outperforms all others.

\begin{figure}[h]
\label{fig:3}
\includegraphics[width=0.48\textwidth]{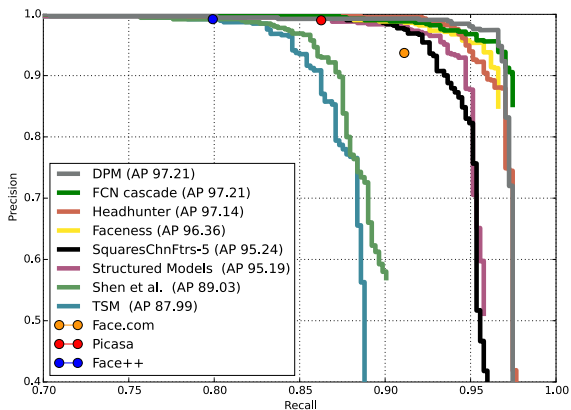}
\centering
\caption{Precision-recall curves on AFW dataset. AP = average precision}
\end{figure}

\textbf{FDDB dataset results.} This dataset is a widely used benchmark for uncontrolled face detection. It consists of 2845 images with 5171 labeled faces collected from  news articles on Yahoo websites. It has a standard evaluation process which we follow to report our detection performance using the toolbox provided with the dataset. FDDB dataset uses ellipse face annotations. To adapt our detection results to elliptical annotations, we uniformly extend the vertical edge in our detection bounding box by 25\%, and then move the center of the box up by 10\% of the vertical edge.

Fig. 4 compares the performance of different face detectors on this dataset using Receiver Operating Characteristic (ROC) curves. The set includes recently published methods: \cite{chen2014joint},\cite{li2015convolutional}, \cite{zhu2012face},  \cite{mathias2014face}, \cite{yang2015facial},  \cite{farfade2015multi}. The results show that the FCN cascade performs better than all of the detectors other than Faceness, but our training requirements are much less demanding than those for Faceness (discussed below). Note that the different detectors have been trained on different training data so the comparison is not just for the classification algorithms. This is unavoidable since only the detector code or detection results are available; we have followed the standard practice in the literature in our comparisons.

As a part-based method, training of Faceness needs attribute labels (25 attributes in total) for 5 facial parts. This is expensive and tedious to do. Besides, Faceness is trained on the CelebFaces dataset with its own attribute labels. It performs quite well on FDDB where the images come from Yahoo news articles, but it does not perform as well on AFW, in which the images come from Flickr. This indicates that Faceness may not be very adaptive. Also, training of 5 attribute CNNs and face detection network takes over 94,000 positive samples, compared with 6000 positive samples in FCN cascade training. All the 6000 positive training samples for FCN come from VGG Face dataset; according to \cite{parkhi2015deep}, all annotations in this dataset come from automatic face detection and partial human filtering. Although we noticed that there is some noise in these annotations, the FCN cascade trained on this dataset still achieves good results.

\begin{figure}[h]
\label{fig:4}
\includegraphics[width=0.47\textwidth,height=63mm]{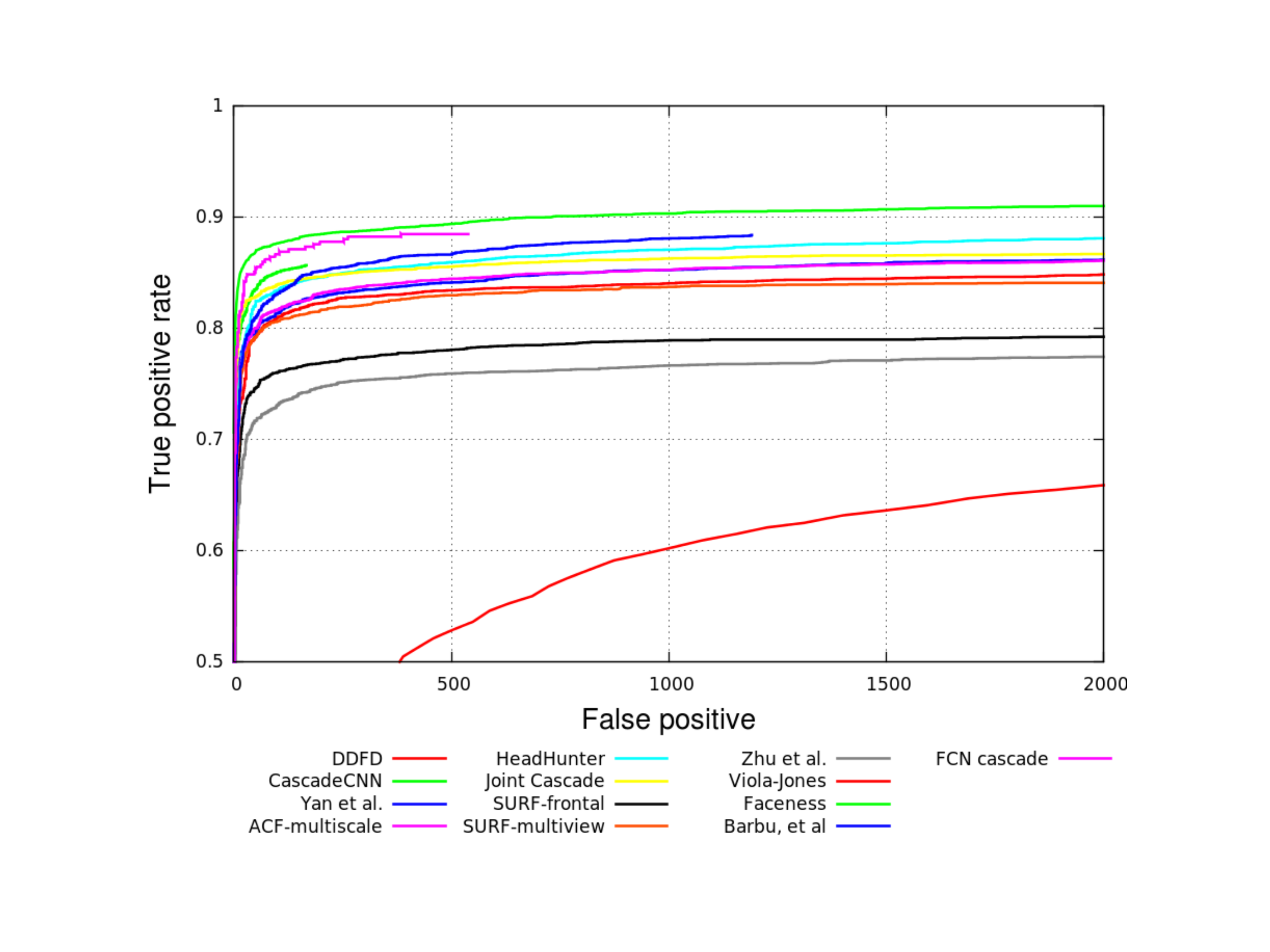}
\centering
\caption{Receiver operating curves on FDDB dataset}
\end{figure}

\textbf{PASCAL faces dataset.} This dataset consists of 851 images with 1335 labeled faces and was collected from the test set of PASCAL person layout dataset, which is a subset of PASCAL VOC \cite{everingham2010pascal}. There are large face appearance and pose variations in this dataset. Fig. 5 shows the precision-recall curves on this dataset.
On this dataset, FCN cascade outperforms all other detectors except Faceness and DPM. Note that this dataset is designed for person layout detection and head annotation is used as face annotation. The cases when the face is occluded are common. So part based methods like DPM and Faceness can have better performance on this dataset.

\begin{figure}[h]
\label{fig:5}
\includegraphics[width=0.46\textwidth,height=58mm]{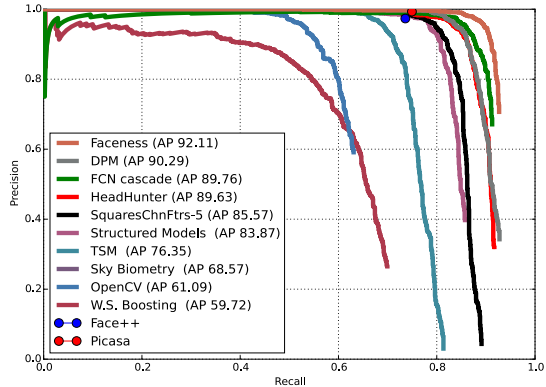}
\centering
\caption{Precision-recall curves on PASCAL faces dataset. }
\end{figure}

\textbf{IJB-A dataset.} We evaluate on this dataset as it has large illumination changes, face pose changes, appearance variation and severe occlusion. It consists of 5397 static images and 20413 frames from videos. The face scales in this dataset range from 10 pixels to up to 2000 pixels. Subjects in this dataset normally are not aware of the the camera so they are not posing for the images. We are aware of only one previous work \cite{ranjan2015deep} that reports results  on this dataset. However, we do not compare to these results as correspondence with authors indicated that the reported results may have errors. Instead, we perform our own tests on four detectors for which code is available to us. We test Viola-Jones, DPM, CNN cascade (an implementation of \cite{li2015convolutional}) and FCN cascade on all 25810 pictures. For DPM, we use pre-trained model by Mathias et al \cite{mathias2014face}.  For Viola Jones, we use the implementaion in OpenCV \cite{bradski2000opencv}. All parameters in these detector implementations are set to default. Fig. 6 shows the precision-recall curves and ROC curves of the four detectors on this dataset. We can see that the four detectors' performance degrade compared to results on AFW, FDDB and PASCAL due to more challenging images in the IJB-A dataset (qualitative detection results are shown in Fig. 7), but the FCN cascade decreases the least in performance. Taking either average precision (AP) in precision-recall curve or area under curve (AUC) in ROC curve as overall detector performance criterion, our methods outperforms the other three detectors. And beat CNN cascade and Viola Jones by a significant margin.

Note that even though DPM has higher precision than FCN cascade at lower recall rates, its AP is significantly lower. And that Run time for FCN cascade is 1.1 secs/frame whereas the DPM, which is the closest in accuracy, is 14.8 secs/frame.

\begin{figure}[h]
\label{fig:6}
\includegraphics[width=0.38\textwidth,height=50mm]{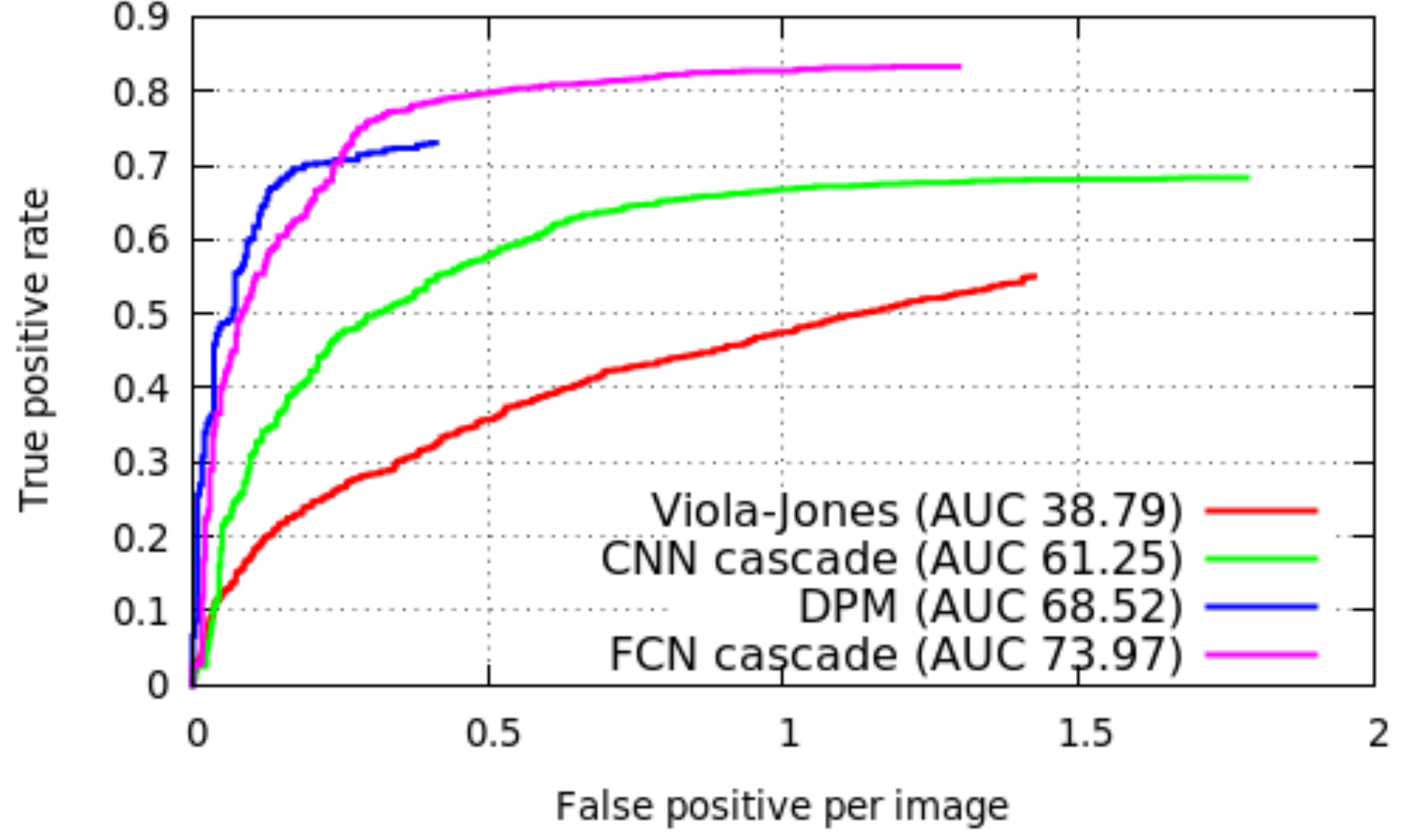}
\centering
\end{figure}
\begin{figure}[h]
\label{fig:7}
\includegraphics[width=0.38\textwidth,height=50mm]{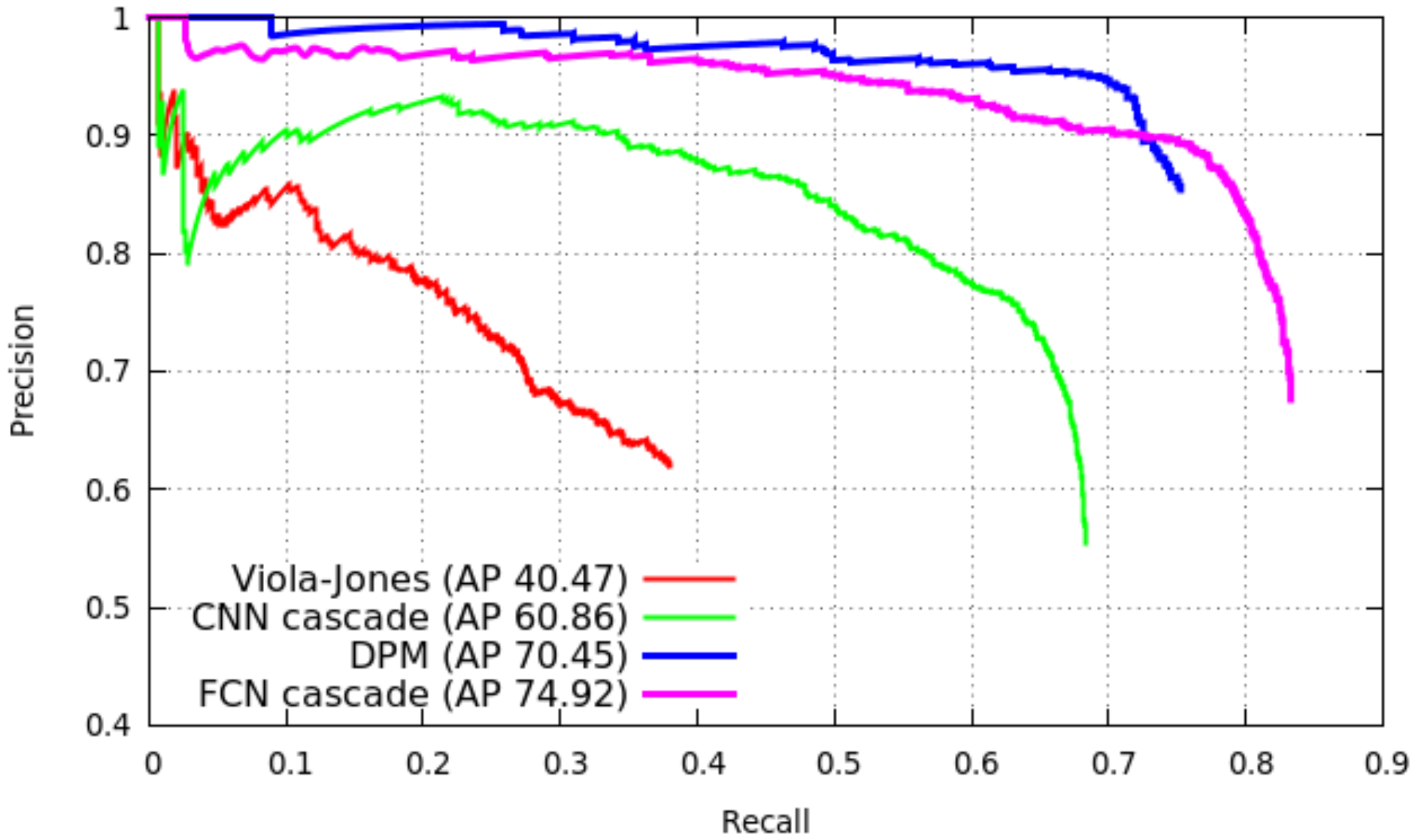}
\centering
\caption{ Performance evaluation on IJB-A dataset. (a) ROC curves. (b) Precision-recall curves}
\end{figure}

\section{Conclusion}
In this study, we explored the application of FCNs and score maps in face detection. We proposed a framework of FCN cascade. We showed that our FCN cascade achieves best result on AFW dataset (tied with DPM), and achieves comparable performance with state-of-art detectors on FDDB and PASCAL face datasets.

The advantages of FCN cascade lie in multiple aspects: 1) FCN cascade deals with multi-scale images. 2) FCN makes it possible to train the network end-to-end. 3) Good performance on different datasets shows that our detector is adaptive.  4) The \begin{figure*}
\label{fig:8}
\includegraphics[width=0.9\textwidth,height=110mm]{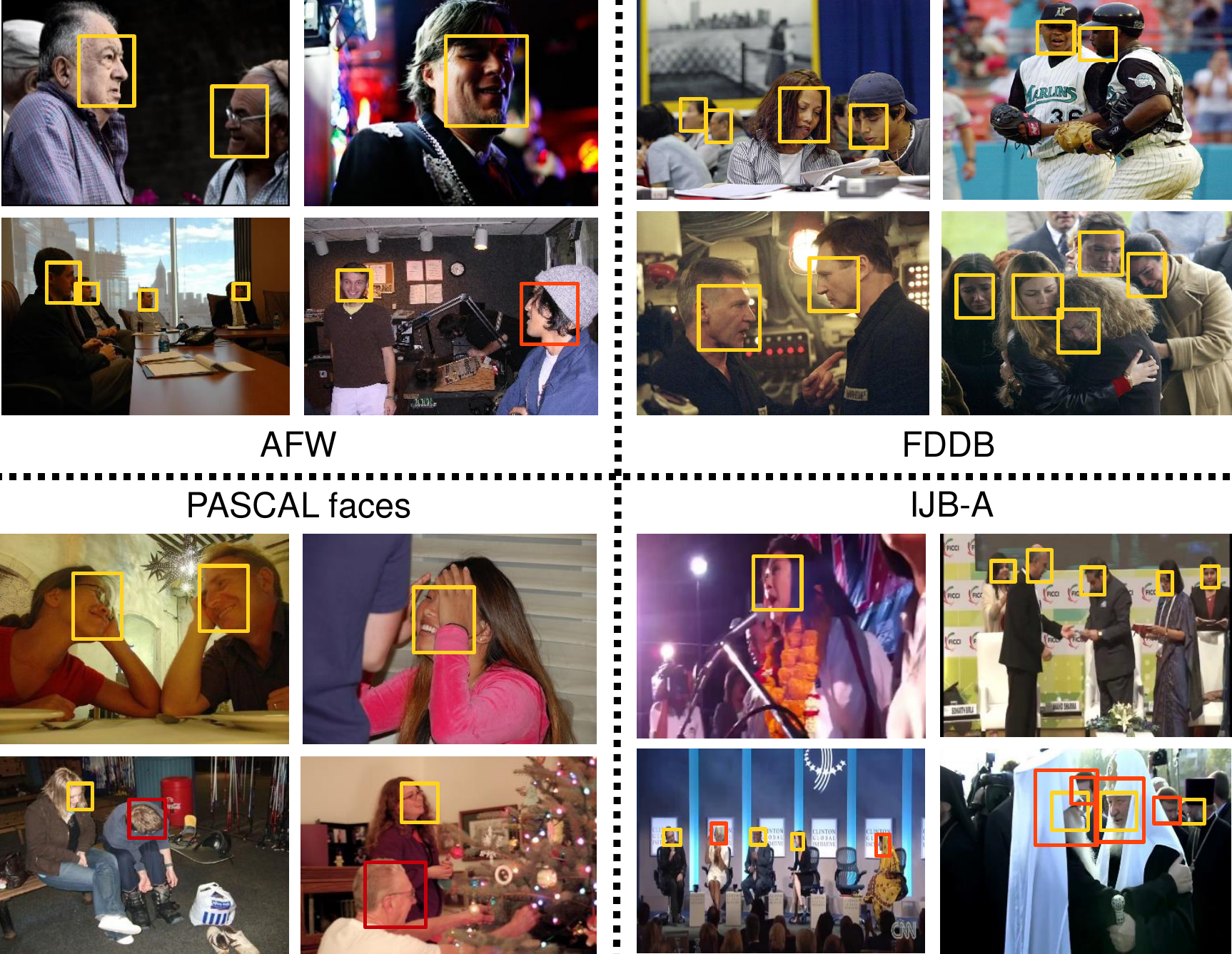}
\centering
\caption{Qualitative detection results on four datasets. Yellow boxes denote true positive detection. Red boxes denote missings.}
\end{figure*}
training of FCN is not highly sensitive to noise in the training samples.

% conference papers do not normally have an appendix

% use section* for acknowledgment
\ifCLASSOPTIONcompsoc
  % The Computer Society usually uses the plural form
  \section*{Acknowledgments}
\else
  % regular IEEE prefers the singular form
  \section*{Acknowledgment}
\fi

This research is based upon work supported, in part, by the Office of the Director of National Intelligence (ODNI), Intelligence Advanced Research Projects Activity (IARPA), via IARPA R\&D Contract No. 2014-14071600011. The views and conclusions contained herein are those of the authors and should not be interpreted as necessarily representing the official policies or endorsements, either expressed or implied, of the ODNI, IARPA, or the U.S. Government. The U.S. Government is authorized to reproduce and distribute reprints for Governmental purposes notwithstanding any copyright annotation thereon.

% trigger a \newpage just before the given reference
% number - used to balance the columns on the last page
% adjust value as needed - may need to be readjusted if
% the document is modified later
%\IEEEtriggeratref{8}
% The "triggered" command can be changed if desired:
%\IEEEtriggercmd{\enlargethispage{-5in}}

% references section

% can use a bibliography generated by BibTeX as a .bbl file
% BibTeX documentation can be easily obtained at:
% http://mirror.ctan.org/biblio/bibtex/contrib/doc/
% The IEEEtran BibTeX style support page is at:
% http://www.michaelshell.org/tex/ieeetran/bibtex/

%\bibliographystyle{IEEEtran}
%\bibliography{IEEEabrv}
%
% <OR> manually copy in the resultant .bbl file
% set second argument of \begin to the number of references
% (used to reserve space for the reference number labels box)

% that's all folks
\end{document}